\documentclass[11pt]{article}

\usepackage[preprint]{acl}

\usepackage{times}
\usepackage{latexsym}

\usepackage[T1]{fontenc}

\usepackage[utf8]{inputenc}

\usepackage{microtype}

\usepackage{inconsolata}

\usepackage{graphicx}

\usepackage{booktabs}

\usepackage{amsmath,amssymb}

\title{Your VLM is not good at counting pushups}

\author{
Shengzhi Li \\
\And
Jiarun Chen \\
\And
Karun Sharma \\
\And
Jiaqi Su \\
\And
Shichao Pei
}
\begin{document}
\maketitle

\begin{abstract}
Large vision-language models (VLMs) can recognize \textit{what} happens in video but fail to count \textit{how many} times. We introduce \textbf{PushupBench}, 446 long-form clips (avg. 36.7s) for evaluating repetition counting. The best frontier model achieves 42.1\% exact accuracy; open-source 4B models score $\sim$6\%, matching supervised baselines. We show that accuracy alone misleads---weaker models exploit the modal count rather than reason temporally. Fine-tuning on counting with just 391 curated samples transfers to general video understanding: MVBench (+2.45), PerceptionTest (+1.23), TVBench (+2.75), suggesting counting is a proxy for broader temporal reasoning. Project page: \url{https://pushupbench.com}
\end{abstract}

% === INTRODUCTION ===
\section{Introduction}

Large vision-language models (VLMs) have achieved remarkable progress in semantic video understanding---correctly identifying that a video shows ``a person doing squats.'' Yet when asked \textit{how many} squats were performed, even frontier models like Gemini 3 Flash struggle, achieving only 42.1\% exact accuracy on our benchmark. This gap between semantic recognition and precise temporal reasoning represents a fundamental limitation: VLMs can describe \textit{what} happens but fail to reliably determine \textit{when} and \textit{how often} \citep{wang2025spacevllm}.

We argue that \textbf{repetition counting}---the task of determining how many times an action occurs in a video---serves as an ideal diagnostic probe for this limitation. Unlike action recognition, which can often be solved from a single frame, counting requires tracking state changes across time, detecting action boundaries, and maintaining a coherent count despite variable speeds, camera motion, and appearance changes. Recent work has shown that VLMs exhibit systematic failures in counting tasks, with performance degrading sharply as object or action count increases due to attention mechanisms that fail to distinctly represent each instance \citep{vlmcantcount2024, vlmcounting2024}.

\paragraph{Prior Work on Repetition Counting.}
Early approaches used Fourier transforms or peak detection on 1D signals derived from frame differences \citep{zhang2020context}, but proved brittle to speed variations and camera motion. Deep learning methods improved robustness: RepNet \citep{dwibedi2020counting} introduced Temporal Self-Similarity Matrices for class-agnostic counting, TransRAC \citep{hu2022transrac} applied Transformers for long-range temporal modeling, and PoseRAC \citep{yao2023poserac} showed competitive results using pose keypoints. % alone.

Crucially, all prior work employed purpose-built architectures predating the video LLM era. Existing benchmarks also have limitations: UCFRep \citep{zhang2020context} and Countix \citep{dwibedi2020counting} contain short clips (8s and 6s avg.), while OVR \citep{dwibedi2024ovr} provides scale (72k videos) but extremely brief durations (2.7s avg.). RepCount \citep{hu2022transrac} offers longer videos (29s avg.) but has not been used to evaluate general-purpose VLMs.

\paragraph{Contributions.}
This paper makes two core contributions:

\textbf{First}, we introduce \textbf{PushupBench}, a benchmark of 446 video clips (avg. 36.7s, range 22--117s) curated from 11 diverse fitness creators across 10 countries. We evaluate both state-of-the-art supervised models and frontier VLMs, finding that all perform poorly: TransRAC achieves only 6.73\% exact match, Gemini 3 Flash reaches 42.1\%, and Qwen3-VL-4B scores 8.9\%. These results confirm that the fine-grained temporal reasoning capability of current LLMs remains fundamentally limited..

\textbf{Second}, we demonstrate that \textbf{training on counting enhances general video understanding}. A Qwen3-VL model fine-tuned on just 391 curated repetition counting samples shows consistent improvements on unrelated benchmarks---MVBench (+2.45), PerceptionTest (+1.23), and TVBench (+2.75)---with gains in temporal-reasoning subtasks while spatial tasks decline, suggesting that counting is not a narrow skill but a proxy for broader temporal reasoning.

% === METHOD ===
\section{Method}

This paper introduces PushupBench, a new benchmark designed to address the identified gaps in the literature and provide a rigorous testbed for evaluating the spatio-temporal reasoning capabilities of modern video understanding models.

% \subsection{Dataset Construction}
Specifically, we constructed PushupBench by manually collecting and segmenting 34 full-length, publicly available exercise videos from YouTube, resulting in a total of 446 individual clips with fine-grained, cycle-level annotations. The data collection was guided by the principle of maximizing diversity across content creators to directly challenge model generalization.

The dataset comprises content from 11 distinct fitness youtuber (see Appendix~\ref{sec:creators} for full details). The selection ensures broad representation across gender, geography, and presentation style. The number of clips segmented from each full-length video ranges from 1 to 25, with a mean of approximately 10.7 clips per video. This structure provides a mix of short,  exercises and long, continuous workout.

% === EXPERIMENTS ===
\section{Experiments}

\subsection{Evaluation Setup}

We evaluate models on PushupBench and four established video understanding benchmarks to assess both counting performance and generalization:

\begin{itemize}
\item \textbf{PushupBench} (Ours): 446 clips (avg. 36.7s) testing repetition counting across diverse fitness content.

\item \textbf{MotionBench} \citep{motionbench2024}: Motion-level perception spanning fine-grained motion, repetition counting, and camera motion---forcing reliance on temporal cues.

\item \textbf{PerceptionTest} \citep{perceptiontest2023}: Probes memory, abstraction, physics, and semantics using scripted videos with multi-modal annotations.

\item \textbf{MVBench} \citep{mvbench2024}: ``Static-to-dynamic'' design converting image tasks to video by injecting temporal requirements (e.g., ``where is X'' $\rightarrow$ ``which direction is X moving'').

\item \textbf{TVBench} \citep{tvbench2024}: Temporally-hard MCQA where questions cannot be answered from single frames or world knowledge alone.
\end{itemize}

For VLM evaluations, we sample at 5 fps and cap at 112 frames; longer videos use uniform spacing to cover the full clip. All frames are resized to 360p with greedy decoding. Claude models are limited to 100 frames by API constraints. We use 10 diverse prompt templates (see Appendix~\ref{sec:prompts}) to avoid overfitting to specific instruction formats. For TransRAC, we use 64 frames at 224$\times$224 following the original protocol.

\paragraph{Evaluation Metrics.}
\label{sec:metrics}
We report three metrics: \textbf{Exact Match} (percentage of predictions equaling ground truth), \textbf{MAE} (mean absolute error), and \textbf{$R^2$} (coefficient of determination). For 27 samples (6.1\%) with ambiguous action boundaries, we accept multiple valid ground truth counts (see Appendix~\ref{sec:annotation}). MAE and $R^2$ are computed excluding predictions with $|\text{error}| > 50$, as extreme outliers from parsing failures or hallucinations can drastically skew these metrics. Prior work reports only MAE and Off-by-One accuracy \citep{dwibedi2020counting, hu2022transrac}; we add $R^2$ because it detects collapse to constant prediction---a failure mode where models achieve non-trivial exact match by exploiting dataset statistics rather than counting. Unlike Pearson correlation, $R^2$ penalizes both bias and scale errors as well.

\subsection{Supervised CV-Baseline}

To establish a non-VLM baseline, we trained TransRAC \citep{hu2022transrac} on our training data mix. TransRAC uses a Video Swin Transformer backbone to extract spatiotemporal features from 64 frames.

\subsection{Main Results: PushupBench Performance}

Table~\ref{tab:countbench} presents zero-shot performance on PushupBench. We evaluate frontier commercial models (Gemini family), open-source VLMs (Qwen3-VL), and a supervised baseline (TransRAC).

\begin{table*}[t]
\centering
\footnotesize
\begin{tabular}{@{}lccc|lccc@{}}
\toprule
\textbf{Model} & \textbf{Exact} & \textbf{MAE} & \textbf{$R^2$} & \textbf{Model} & \textbf{Exact} & \textbf{MAE} & \textbf{$R^2$} \\
\midrule
\multicolumn{4}{l|}{\textit{Commercial VLMs (Gemini @ 5fps)}} & \multicolumn{4}{l}{\textit{Open-Source VLMs (Qwen3-VL @ 5fps)}} \\
Gemini 3 Flash & \textbf{42.1} & \textbf{2.9} & \textbf{0.82} & Qwen3-VL-4B-Instruct & 8.9 & 8.2 & $-$0.21 \\
Gemini 3 Pro & 39.8 & 3.6 & 0.70 & Qwen3-VL-4B-Thinking & 8.2 & 8.9 & $-$0.34 \\
Gemini 2.5 Pro & 29.9 & 5.7 & 0.63 & \quad\textbf{+ DAPO (Ours)} & \underline{14.5} & \underline{5.7} & \underline{0.38} \\
Gemini 2.5 Flash & 10.9 & 7.7 & 0.49 & Qwen3-VL-8B-Instruct & 7.2 & 7.9 & $-$0.11 \\
Gemini 2.0 Flash & 8.9 & 7.8 & 0.25 & Qwen3-VL-8B-Thinking & 5.3 & 9.5 & $-$0.29 \\
\midrule
\multicolumn{4}{l|}{\textit{Commercial VLMs (Other)}} & \multicolumn{4}{l}{\textit{Open-Source VLMs (Qwen3-VL, cont.)}} \\
GPT-5 & 10.9 & 7.6 & 0.01 & Qwen3-VL-30B-A3B-Instruct & 7.6 & 12.6 & $-$0.06 \\
Claude Sonnet 4.5 & 9.5 & 9.0 & 0.00 & Qwen3-VL-30B-A3B-Thinking & 9.6 & 8.8 & 0.08 \\
Claude Opus 4.5 & 4.9 & 9.7 & 0.09 & Qwen3-VL-32B-Instruct & 11.8 & 7.7 & 0.36 \\
LLaMA-4-Maverick & 1.0 & 10.2 & $-$0.15 & Qwen3-VL-32B-Thinking & 9.9 & 9.1 & 0.35 \\
 & & & & Qwen3-VL-235B-Instruct & 8.6 & 7.1 & 0.17 \\
\midrule
\multicolumn{4}{l|}{\textit{Open-Source VLMs (Qwen2.5-VL @ 5fps)}} & \multicolumn{4}{l}{\textit{Baselines}} \\
Qwen2.5-VL-3B & 5.6 & 8.9 & $-$0.86 & TransRAC (trained on same data) & 6.7 & 9.1 & $-$0.18 \\
Qwen2.5-VL-7B & 9.2 & 7.8 & $-$0.40 & Const.\ (mode=10) & 9.9 & 8.5 & $-$0.21 \\
 & & & & Const.\ (mean=17) & 2.9 & 9.2 & 0.00 \\
\bottomrule
\end{tabular}
\caption{PushupBench results. $R^2$ (coefficient of determination) measures variance explained beyond the mean baseline; positive $R^2$ indicates predictions that scale with ground truth; negative values indicate constant or random output }
\label{tab:countbench}
\end{table*}

\subsection{Generalization to Video Understanding}

A key finding is that training on repetition counting transfers to general video understanding tasks. Table~\ref{tab:generalization} shows performance on four established benchmarks before and after fine-tuning.

\begin{table}[tb]
\centering
\small
\begin{tabular}{lcccc}
\toprule
\textbf{Model} & \textbf{Motion} & \textbf{Percept.} & \textbf{MV} & \textbf{TV} \\
\midrule
Base & 58.2 & 72.57 & 65.75 & 53.16 \\
+ DAPO & \textbf{58.7} & \textbf{73.80} & \textbf{68.20} & \textbf{55.91} \\
\midrule
$\Delta$ & +0.5 & +1.23 & +2.45 & +2.75 \\
\bottomrule
\end{tabular}
\caption{Generalization results (Qwen3-VL-4B-Thinking). Fine-tuning on counting with 391 curated samples improves all four video benchmarks, with the largest gain on TVBench (+2.75).}
\label{tab:generalization}
\end{table}

\paragraph{Counting generalizes.} Category-level analysis reveals improvements concentrate in tasks requiring temporal state tracking. The top-5 improving subtasks across MVBench and TVBench are all temporal-reasoning tasks: TVBench \textit{moving\_direction} (+35.3), MVBench \textit{counterfactual\_inference} (+13.5), TVBench \textit{action\_sequence} (+12.8), MVBench \textit{moving\_count} (+9.5), and TVBench \textit{action\_count} (+7.7). Non-temporal tasks are flat or decline. Notably, \textbf{spatial} object counting (TVBench \textit{object\_count}) \textit{drops} by 17.6 points while \textbf{temporal} action counting improves by +7.7---the opposite of what generic video training would produce. Our 391 fitness-only samples are $<$0.003\% of Qwen3-VL's 16M-example post-training budget; if generic video exposure sufficed, the existing training would have already produced these gains.

\subsection{Fine-Tuning Setup}

We initially curated 968 training samples from three public repetition counting datasets (OVR-Kinetics, RepCount, and our own custom workout annotations). Post-submission analysis revealed that training instability stemmed primarily from \textbf{data quality, not reward design or model limitations}. We identified two issues:

\paragraph{Temporally unresolvable samples.} OVR-Kinetics---our largest training source (540 of 968 samples)---contains videos where repetitions are too fast for 5fps sampling. At 5fps, 46.4\% of OVR samples have $<$3 frames per repetition (mean 3.3 frames/rep), compared to 0\% for our custom annotations (mean 16.3 frames/rep) and 2.3\% for RepCount (mean 11.0 frames/rep). Training on these samples injects noise into the reward signal.

\paragraph{Mode collapse at GT=10.} Samples with ground-truth count of exactly 10 comprised 10.6\% of training data, creating a mode-collapse attractor where the model achieves disproportionate reward by always predicting ``10.''

We removed OVR entirely and all GT=10 samples, reducing training from 968 to \textbf{391 samples} ($-$60\%). Despite using 60\% fewer samples, all metrics improved (Table~\ref{tab:countbench}: exact match 5.9\% $\rightarrow$ 14.5\%, $R^2$ 0.23 $\rightarrow$ 0.38). This shows that for RL video training, data quality matters more than quantity. See Appendix~\ref{sec:training_details} for full dataset statistics.

We fine-tune Qwen3-VL-4B-Thinking using VERL \citep{verl2024} with the DAPO algorithm \citep{grpo2024}, sampling 112 frames at 5 fps. We use a scale-invariant linear reward $r = \max(0, 1 - |\hat{c} - c_{\text{GT}}|/c_{\text{GT}})$ that provides smooth gradients even when predictions are not exact---critical for GRPO which requires within-group reward variance to compute meaningful advantages (see Appendix~\ref{sec:reward_design}). Full hyperparameters are provided in Appendix~\ref{sec:training_details}.

% === DISCUSSION ===
\section{Discussion}

\subsection{Frame Rate Requirements}
\label{sec:framerate}

By the Nyquist-Shannon sampling theorem, we need at least twice the motion's frequency as our sampling frequency to capture all information needed to detect repetitions. For a video containing $N$ repetitions, we require at least $2N$ frames.We validate this empirically with an FPS ablation study (Appendix~\ref{sec:fps_ablation}): $R^2$ improves from 0.61 at 1 fps to 0.82 at 5 fps as more videos exceed the Nyquist threshold.

\subsection{Error Analysis: $R^2$ Reveals Counting Failure}

\begin{figure}[t]
\centering
\includegraphics[width=\columnwidth]{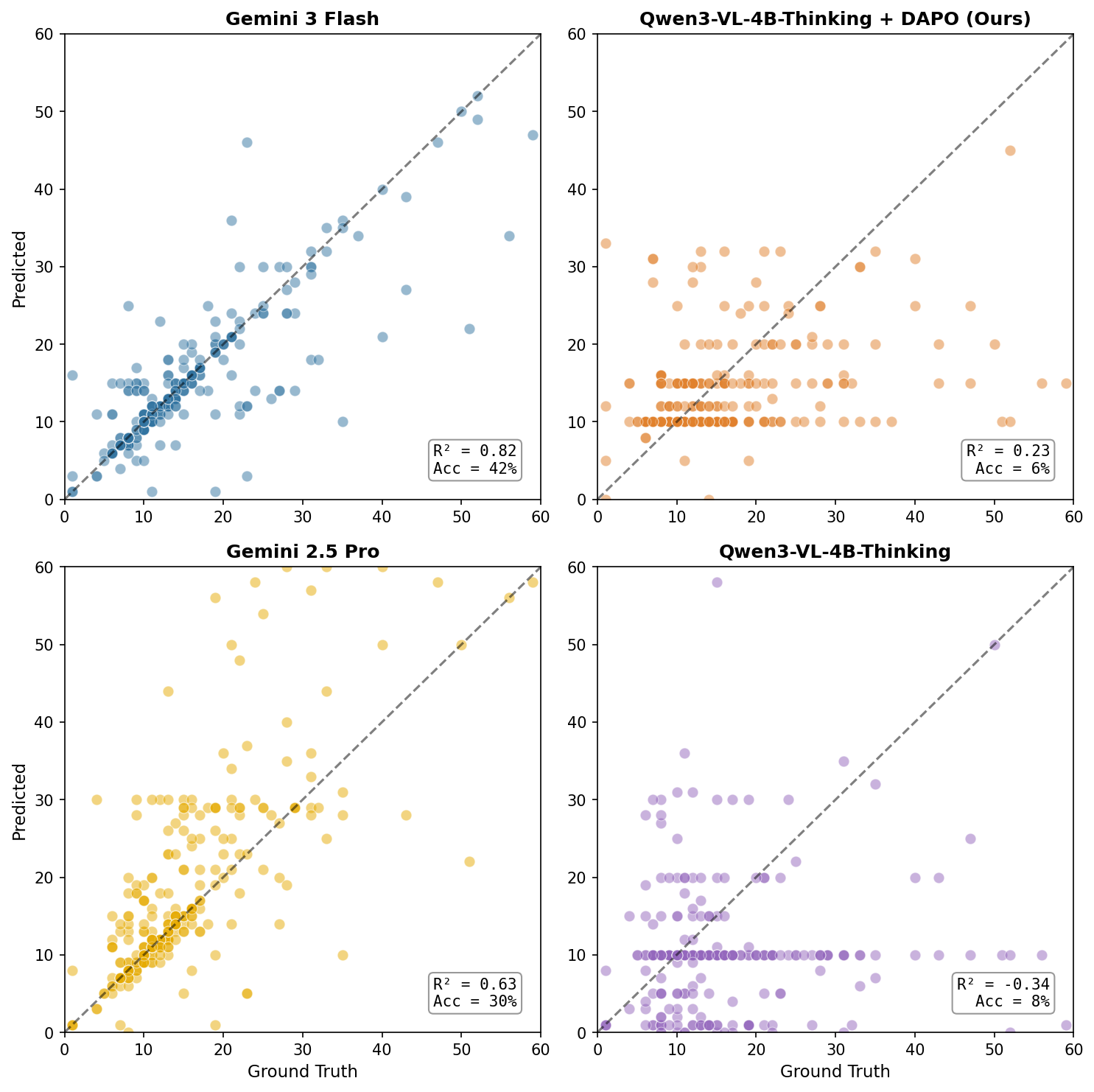}
\caption{Predicted vs. ground truth repetition counts for four models. Diagonal line = perfect prediction. Gemini 3 Flash shows strong fit ($R^2$=0.82); base Qwen3-VL-4B-Thinking collapses to constant predictions ($R^2$=$-$0.34). Fine-tuning with DAPO on 391 curated samples improves both $R^2$ ($-$0.34 $\rightarrow$ 0.38) and exact match (8.2\% $\rightarrow$ 14.5\%), demonstrating genuine counting improvement.}
\label{fig:scatter}
\end{figure}

\begin{figure}[t]
\centering
\includegraphics[width=\columnwidth]{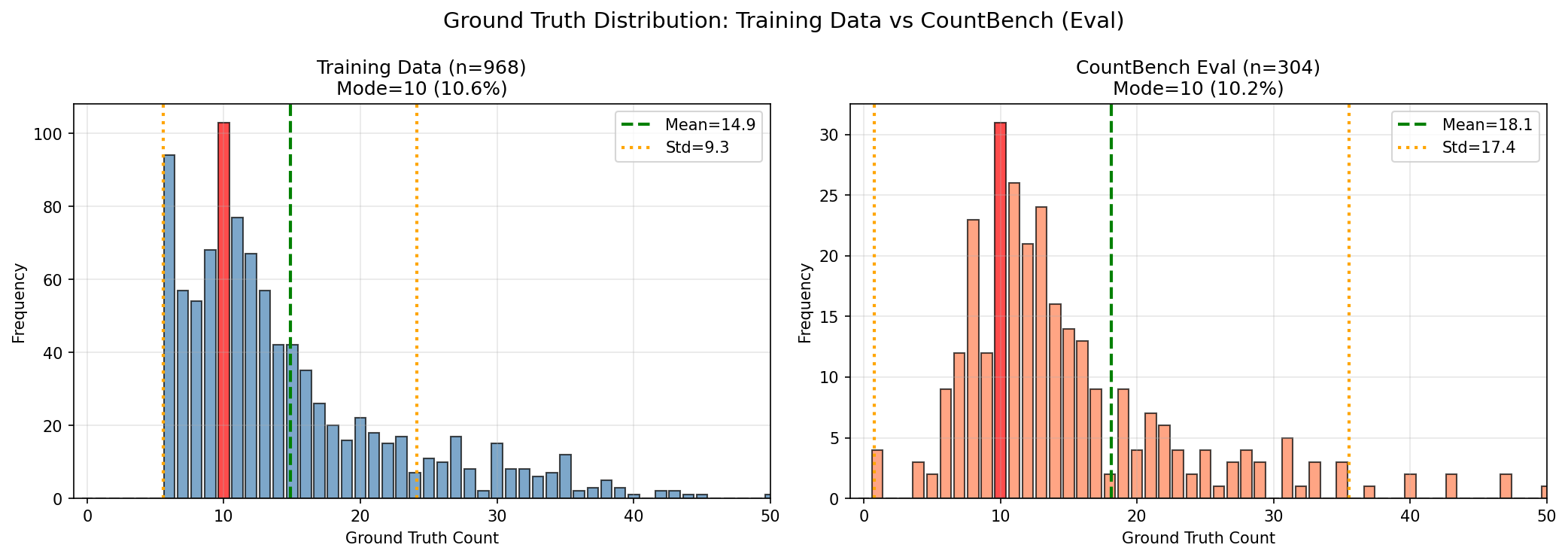}
\caption{Ground truth distribution in training data (left) and PushupBench (right). Both datasets share mode=10, reflecting a natural human preference for 10-rep workout sets. This explains why weaker models collapse to predicting ``10'' regardless of input (Figure~\ref{fig:scatter}, bottom-right): outputting the dataset mode achieves $\sim$10\% exact match without any temporal reasoning.}
\label{fig:gt_distribution}
\end{figure}

Figure~\ref{fig:scatter} visualizes prediction vs. ground truth for four models, revealing distinct failure modes. \textbf{Gemini 3 Flash} ($R^2$=0.82, 42\% exact) shows strong fit---predictions track ground truth with reasonable accuracy. \textbf{Gemini 2.5 Pro} ($R^2$=0.63, 30\% exact) shows good correlation but higher variance. In contrast, the open-source models exhibit qualitatively different failures: \textbf{Qwen3-VL-4B-Thinking} ($R^2$=$-$0.34, 8\% exact) collapses to constant predictions around 10 regardless of ground truth---achieving non-trivial exact match by exploiting dataset statistics rather than counting. As shown in Figure~\ref{fig:gt_distribution}, this behavior reflects a natural property of fitness data: people commonly perform exercises in sets of 10 reps, making 10 the modal count in both training and evaluation data (see Appendix~\ref{sec:training_details} for training dynamics).

Critically, fine-tuning transforms this failure mode: \textbf{Qwen3-VL-4B + DAPO} ($R^2$=0.38, 14.5\% exact) shows predictions that scale with ground truth, indicating genuine counting behavior. With curated training data (391 samples, see Section 3.4), both $R^2$ and exact match improve ($R^2$: $-$0.34 $\rightarrow$ 0.38; exact: 8.2\% $\rightarrow$ 14.5\%). This illustrates why $R^2$ is essential: it distinguishes models that count from those that exploit statistics.

\subsection{Reward Hacking and Data Curation}

During RL fine-tuning, we identified two reward hacking patterns requiring mitigation.

\paragraph{Mode Collapse.} Since GT=10 is the dataset mode (Figure~\ref{fig:gt_distribution}), always predicting ``10'' yields non-zero expected reward without any counting. We observed checkpoints collapse to this behavior; reward shaping mitigated this (Appendix~\ref{sec:reward_design}).

\paragraph{On-Screen Text Exploitation.} Models read on-screen counters and timers rather than counting motion. Analysis of 23k rollouts from an early training run revealed a critical pattern: genuine counting accuracy collapses above GT=10 (31\% for GT 1--5, dropping to $\sim$0\% for GT$>$15), while hacking attempts increase with GT count (1--2\% for GT$\leq$10 vs.\ 35\% for GT$>$40). This means for high-count samples, nearly all non-zero reward gradients came from hacking rollouts rather than genuine counting (Table~\ref{tab:hacking_prevalence}). Based on this analysis, we excluded GT$<$6 samples (where the model already performs reasonably) and removed videos with on-screen clues. For the benchmark, we manually edited out overlay artifacts (Appendix~\ref{sec:reward_hacking}).

% === CONCLUSION ===
\section{Conclusion}

We introduced PushupBench, a challenging benchmark for evaluating fine-grained temporal reasoning in video understanding models. Our experiments reveal that both frontier VLMs and supervised approaches struggle with repetition counting, achieving at best 42.1\% exact accuracy. Critically, we demonstrate that training on just 391 curated counting samples transfers to general video reasoning---with gains in temporal-reasoning subtasks across MVBench, PerceptionTest, and TVBench, while spatial tasks decline. This suggests repetition counting is an effective proxy task for learning temporal reasoning in vision-language models.

% === LIMITATIONS ===
\section*{Limitations}

Our benchmark focuses exclusively on fitness videos, which may not capture the full diversity of repetitive actions in other domains (e.g., industrial, sports, daily activities). Additionally, we only tuned one open-source model family (Qwen3-VL) due to computational constraints.

Our RL fine-tuning used 391 curated samples after removing temporally unresolvable OVR-Kinetics data and mode-collapse-inducing GT=10 samples (reduced from an initial 968). The 391 clean samples outperformed the original 968 noisy ones on all metrics, showing that quality matters more than scale for this task. While we mitigated reward hacking from frame count inference and on-screen text exploitation, and experimented with multiple reward functions (Appendix~\ref{sec:reward_design}), we only tuned one model family (Qwen3-VL) due to computational constraints. Extending to other VLM families is important future work; concurrent work such as Time-R1 has shown that similar RL post-training can transfer across multiple backbones.

% === REFERENCES ===
\bibliography{custom}

\appendix

\section{Prompt Templates}
\label{sec:prompts}

We created 10 different instruction prompt templates with varying degrees of step-by-step instruction or direct probing. The benchmark data and training data are randomly assigned to one of the ten prompts. We do not assume a specific chain-of-thought format is better than another; this diversity provides the most realistic measure of how models perform in the wild. All prompts request answers in \texttt{\textbackslash boxed\{\}} format for consistent extraction.

\begin{table*}[t]
\centering
\small
\begin{tabular}{clp{10cm}}
\toprule
\textbf{\#} & \textbf{Style} & \textbf{Prompt Template} \\
\midrule
1 & Step-by-step & \texttt{<video>} Watch this video carefully and count the number of repetitions of the action: ``\{description\}''. Think step by step: 1. Identify the repeating action in the video 2. Count each complete repetition carefully 3. Provide your final count. How many repetitions are there? Put your answer in \textbackslash boxed\{\}. \\
\midrule
2 & Direct Question & \texttt{<video>} Count how many times the action ``\{description\}'' is repeated in this video. Watch carefully and count each complete repetition. Put your final answer in \textbackslash boxed\{\}. \\
\midrule
3 & Observation-focused & \texttt{<video>} Observe this video showing ``\{description\}''. Your task is to count the total number of repetitions performed. Pay attention to when each repetition starts and ends. What is the count? Answer in \textbackslash boxed\{\}. \\
\midrule
4 & Analytical & \texttt{<video>} Analyze this video and determine how many times ``\{description\}'' is performed. Consider: When does each repetition begin and end? How many complete cycles are shown? Provide the total count in \textbackslash boxed\{\}. \\
\midrule
5 & Concise & \texttt{<video>} Video shows: ``\{description\}'' Count the repetitions. Put the number in \textbackslash boxed\{\}. \\
\midrule
6 & Instructional & \texttt{<video>} You are watching a video of someone performing ``\{description\}''. Please count how many complete repetitions are shown from start to finish. Enter your count in \textbackslash boxed\{\}. \\
\midrule
7 & Task-oriented & \texttt{<video>} Task: Count repetitions. Action: ``\{description\}''. Watch the entire video and count how many times the action is completed. Total repetitions: \textbackslash boxed\{\} \\
\midrule
8 & Question-Answer & \texttt{<video>} Question: How many repetitions of ``\{description\}'' are performed in this video? Watch carefully, then provide your answer in \textbackslash boxed\{\}. \\
\midrule
9 & Detailed Observation & \texttt{<video>} In this video, a person is performing ``\{description\}''. Count each time the action is fully completed. A repetition counts only when the full motion cycle is finished. How many repetitions? \textbackslash boxed\{\} \\
\midrule
10 & Simple \& Direct & \texttt{<video>} ``\{description\}'' - count the reps. Answer in \textbackslash boxed\{\}. \\
\bottomrule
\end{tabular}
\caption{The 10 prompt templates used for training and evaluation. Templates vary from detailed step-by-step instructions to concise direct queries, ensuring models are not overfit to a specific instruction format.}
\label{tab:prompts}
\end{table*}

\section{Annotation Process}
\label{sec:annotation}

Two authors of this paper independently collected and annotated all exercise videos; no external annotators, crowdworkers, or paid participants were used. As paper authors conducting their own research, no compensation was provided for annotation work. The annotation process followed a two-phase protocol:

\paragraph{Phase 1: Independent Annotation.} Each author independently watched full workout videos and recorded annotations following agreed-upon guidelines. For each exercise segment, annotators recorded: (1) start time, (2) end time, (3) exercise name/description, and (4) repetition count.

\paragraph{Phase 2: Cross-Validation.} One author reviewed the other's annotations, checking temporal boundaries and counts. For ambiguous cases where the reviewer could not determine a definitive count (e.g., partial repetitions, unclear motion boundaries), we marked the exercise as ``fuzzy'' and recorded multiple acceptable ground truth values.

\paragraph{Annotation Schema.} Each video's annotations are stored as JSON with the following structure:

\begin{verbatim}
{
  "video_id": "6770817f4ecee7...",
  "name": "side plank reach",
  "start_time": "06:14:00",
  "end_time": "06:44:00",
  "count": [11, 10, 12],
  "fuzzy_action": true
}
\end{verbatim}

For fuzzy actions, the \texttt{count} field contains multiple acceptable values (e.g., [11, 10, 12] indicates counts of 10, 11, or 12 are all considered correct). Of the 446 clips in PushupBench, 27 (6.1\%) are marked as fuzzy actions.

\section{Exercise Diversity}
\label{sec:exercises}

\begin{figure}[t]
\centering
\includegraphics[width=\columnwidth]{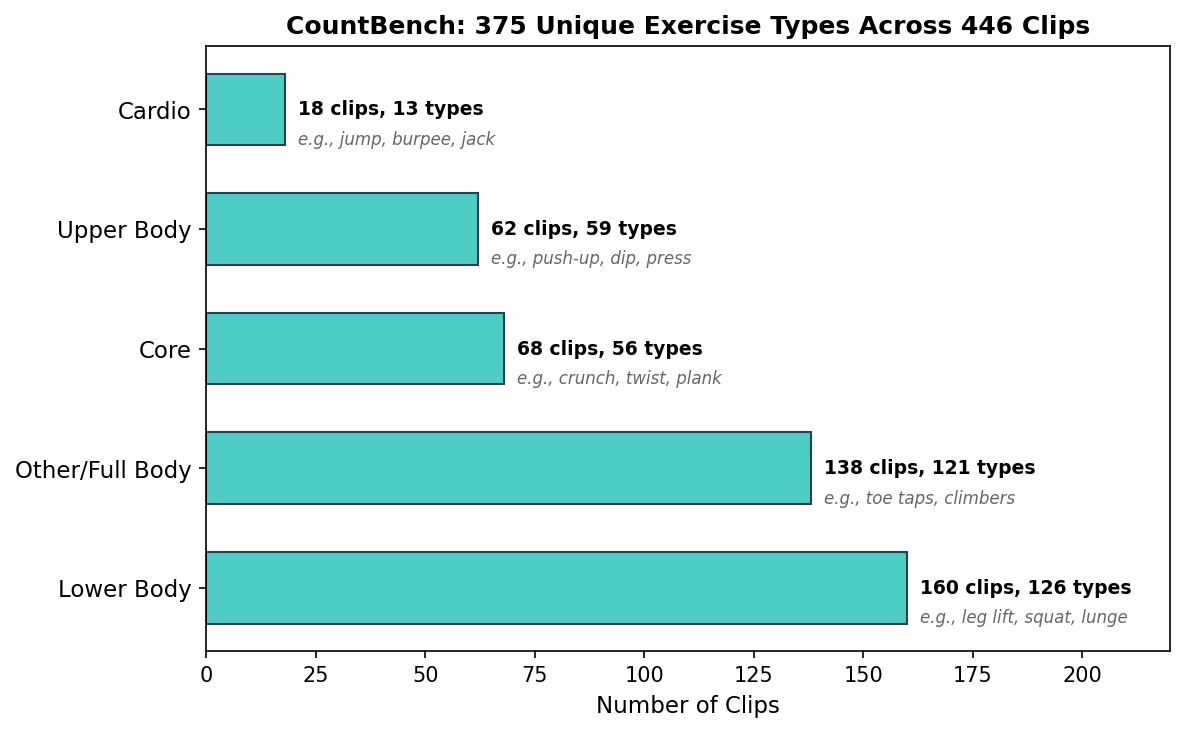}
\caption{Exercise type diversity in PushupBench. The dataset spans 375 unique exercise types across 446 clips, including leg lifts, squats, lunges, crunches, planks, and jumping jacks. This high diversity (84\% unique) ensures evaluation of generalization rather than memorization.}
\label{fig:exercise_diversity}
\end{figure}

PushupBench contains 446 clips spanning \textbf{375 unique exercise types}---an 84\% uniqueness ratio that ensures models cannot rely on memorizing specific action patterns. Figure~\ref{fig:exercise_diversity} shows the distribution across five body-part categories.

\section{Content Creator Diversity}
\label{sec:creators}

We constructed PushupBench by manually collecting and segmenting 34 full-length, publicly available exercise videos from YouTube, resulting in 446 individual clips with fine-grained, cycle-level annotations. The data collection was guided by the principle of maximizing diversity across content creators to directly challenge model generalization. Table~\ref{tab:creators} shows the full list of content creators included in the benchmark.

\begin{table}[t]
\centering
\small
\begin{tabular}{@{}llcr@{}}
\toprule
\textbf{Creator} & \textbf{Country} & \textbf{G} & \textbf{\#} \\
\midrule
Pamela Reif & Germany & F & 6 \\
Chloe Ting & Australia & F & 1 \\
Caroline Girvan & N. Ireland & F & 1 \\
Growingannanas & Austria & F & 1 \\
Eylem Abaci & Germany & F & 8 \\
Chris Heria & USA & M & 1 \\
MIZI & Korea/Malaysia & F & 1 \\
Toned w/ Alexandra & Sweden/Spain & F & 9 \\
Shirlyn Kim & South Korea & F & 2 \\
Lucy Wyndham-Read & UK & F & 1 \\
Oliver Sjostrom & Sweden & M & 3 \\
\midrule
\textbf{Total} & \textbf{10 Regions} & \textbf{9/2} & \textbf{34} \\
\bottomrule
\end{tabular}
\caption{Content creator diversity in PushupBench. G=Gender, \#=Videos. The dataset spans 11 creators across 10 countries.}
\label{tab:creators}
\end{table}

\section{Reward Hacking Analysis}
\label{sec:reward_hacking}

RL training is vulnerable to spurious correlations where models find shortcuts that maximize reward without learning the intended behavior. During fine-tuning on repetition counting, we identified and addressed three distinct forms of reward hacking.

\paragraph{Mode Collapse.}
Since GT=10 is the dataset mode (Figure~\ref{fig:gt_distribution}), always predicting ``10'' yields non-zero expected reward. With our linear percentage reward $r = \max(0, 1 - |\hat{c} - c_{\text{GT}}|/c_{\text{GT}})$, predicting 10 for GT values 6--15 still yields $r > 0.3$. Given that $\sim$40\% of training samples fall in this range, mode collapse becomes a profitable strategy. We observed checkpoints achieve 11\% exact match while predicting ``10'' for 99\% of samples. Reward function analysis in Appendix~\ref{sec:reward_design} shows how different reward designs affect collapse vulnerability.

\paragraph{Frame Count Exploitation.}
In early experiments, we sampled exactly $2N$ frames for videos with $N$ repetitions, following the Nyquist minimum. While training reward increased steadily, benchmark performance did not improve. Analysis revealed the model had learned a shortcut: output half the frame count rather than counting repetitions. We addressed this by using a fixed sampling rate (5 fps) regardless of repetition count, breaking the deterministic relationship between input frames and ground truth.

\begin{figure}[t]
\centering
\includegraphics[width=0.48\columnwidth]{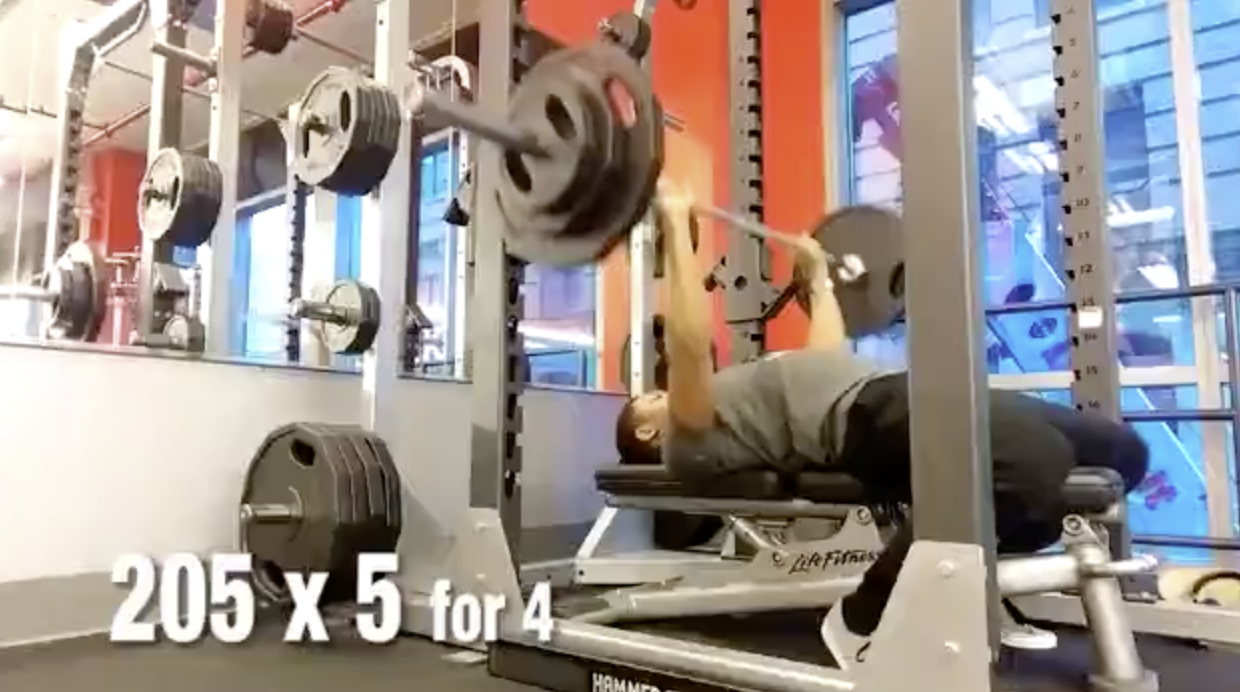}
\includegraphics[width=0.48\columnwidth]{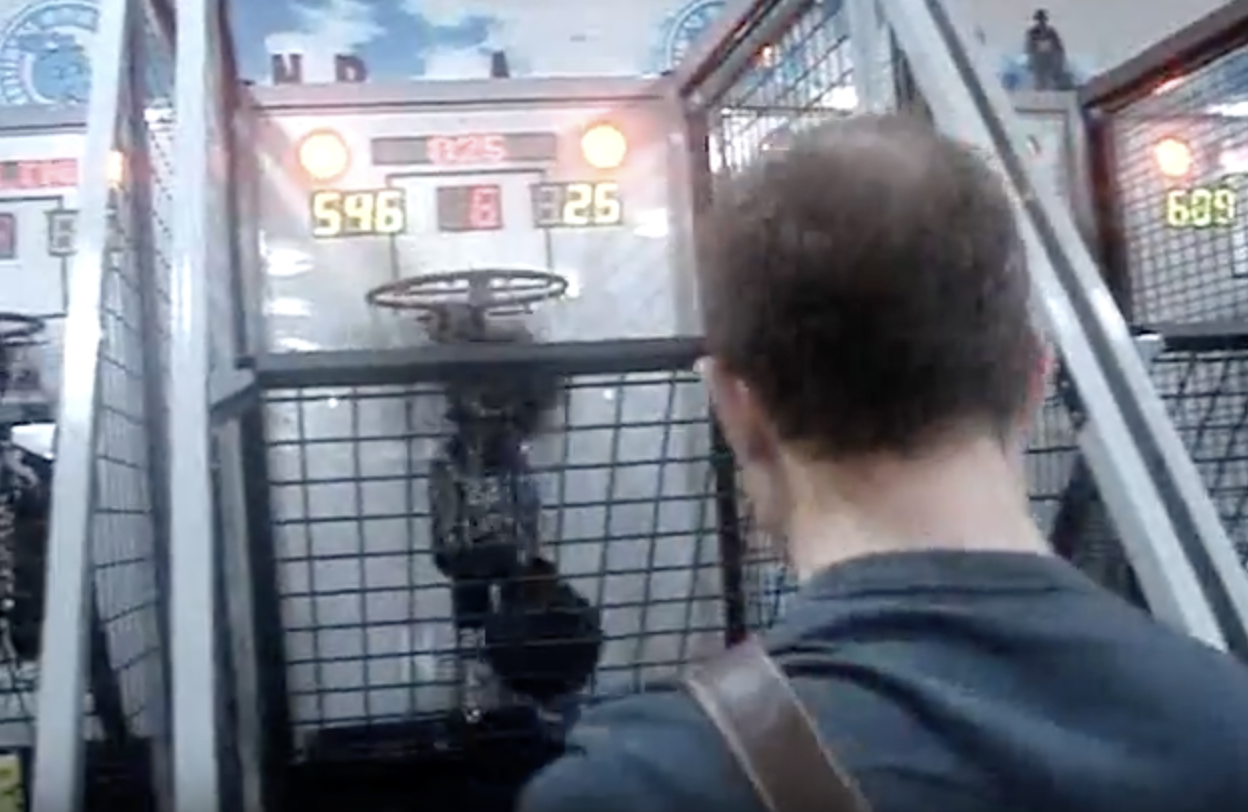}
\caption{Examples of on-screen clues enabling reward hacking. \textbf{Left}: Workout parameters (``205 x 5 for 4'') revealing rep count. \textbf{Right}: Digital scoreboard numbers that could be misread as action counts.}
\label{fig:onscreen_examples}
\end{figure}

\paragraph{On-Screen Counter Exploitation.}
Many fitness videos include on-screen repetition counters, timers, or workout parameter overlays as visual aids. VLMs preferentially attend to these text overlays rather than the underlying motion, achieving high accuracy by reading displayed numbers---OCR rather than temporal reasoning. Figure~\ref{fig:onscreen_examples} shows examples of on-screen clues that enable this shortcut.

\begin{figure}[t]
\centering
\includegraphics[width=0.48\columnwidth]{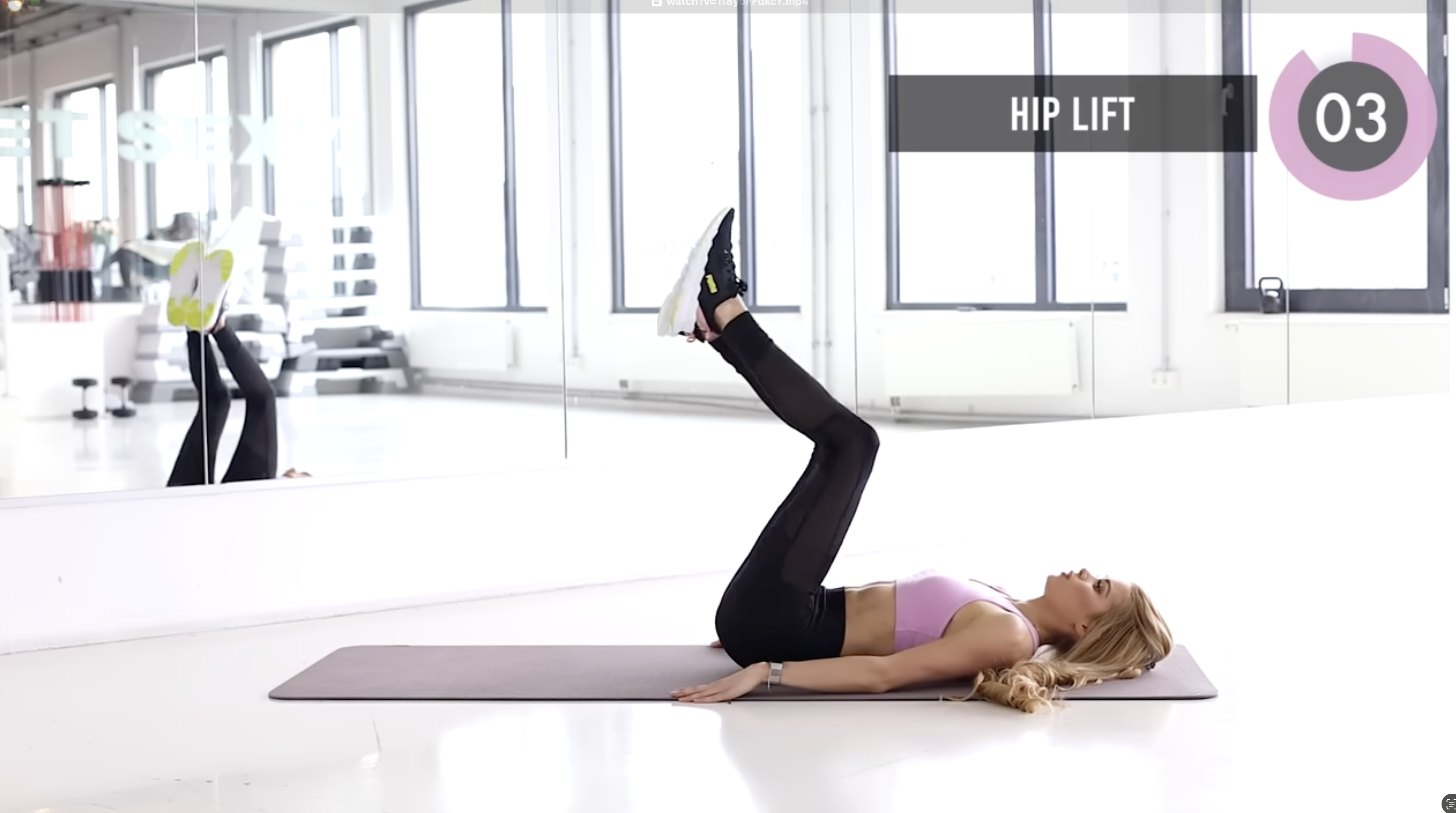}
\includegraphics[width=0.48\columnwidth]{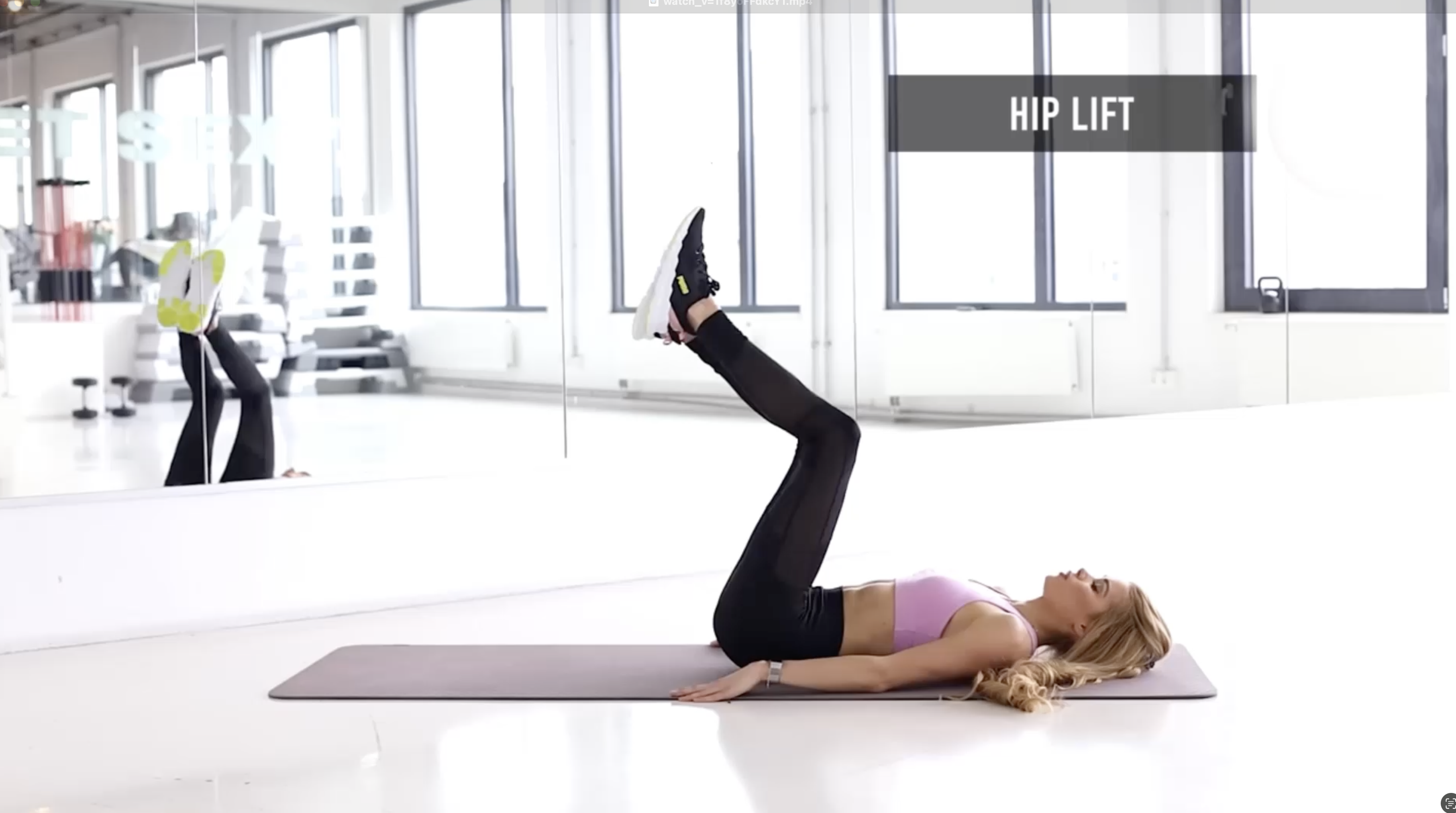}
\caption{On-screen counter removal for benchmark curation. \textbf{Left}: Original frame with rep counter ``03'' visible. \textbf{Right}: Same frame with counter edited out to prevent text-based shortcuts.}
\label{fig:counter}
\end{figure}

For our benchmark, we addressed this by either removing problematic videos entirely or manually editing out textual artifacts. Figure~\ref{fig:counter} shows an example of such editing: the original frame contains a visible rep counter ``03'' which was removed to ensure evaluation measures genuine counting ability.

\paragraph{Prevalence Analysis.}
To understand the scope of this problem, we analyzed rollout data from an early training experiment (23,040 samples) before data curation. Table~\ref{tab:hacking_prevalence} shows reward hacking prevalence across ground truth count buckets. Hacking attempts---identified by model outputs referencing timers, counters, or on-screen text---are rare overall (3.4\%) but increase dramatically with ground truth count: 1--2\% for GT 1--10 versus 29\% for GT 51+. This pattern reflects that higher-count videos are more likely to contain on-screen counters. Critically, genuine counting accuracy collapses above GT=10 (31.1\% for GT 1--5, dropping to near 0\% for GT 16+), suggesting models cannot reliably count beyond $\sim$10 repetitions without resorting to on-screen text.

\begin{table}[t]
\centering
\small
\begin{tabular}{lcccc}
\toprule
\textbf{GT Range} & \textbf{Total} & \textbf{Hack\%} & \textbf{Hack Acc} & \textbf{Count Acc} \\
\midrule
1--5 & 9,364 & 1.3\% & 27.6\% & 31.1\% \\
6--10 & 6,496 & 2.2\% & 33.3\% & 9.0\% \\
11--15 & 2,740 & 4.7\% & 7.7\% & 0.8\% \\
16--20 & 1,640 & 6.3\% & 17.5\% & 0.0\% \\
21--30 & 1,868 & 5.5\% & 22.3\% & 0.3\% \\
31--50 & 796 & 18.0\% & 15.4\% & 0.4\% \\
51+ & 136 & 29.4\% & 30.0\% & 2.2\% \\
\midrule
\textbf{Total} & 23,040 & 3.4\% & 21.1\% & 15.6\% \\
\bottomrule
\end{tabular}
\caption{Reward hacking prevalence by ground truth count from early training runs rollouts. Hacking attempts increase with GT (1--2\% for low counts vs 29\% for 51+), while genuine counting accuracy collapses above GT=10, and only graident were coming from hacking rollouts. Based on this analysis, problematic samples were removed from subsequent training and 1-5G GT range is also excluded.}
\label{tab:hacking_prevalence}
\end{table}

\paragraph{Detection Methodology.}
We detected reward hacking by analyzing model outputs for patterns indicating reliance on on-screen text rather than motion counting:
\begin{itemize}
\item \textbf{Timers}: References to countdown timers being mistaken for repetitions (e.g., ``timer starts from'', ``one rep per second'')
\item \textbf{Counters}: Mentions of on-screen rep counters (e.g., ``counter shows'', ``numbers displayed on screen'')
\item \textbf{Subtitles}: Text overlays containing numbers (e.g., ``text says 10'', ``on-screen text'')
\end{itemize}

\paragraph{Training Data Curation.}
Based on this analysis, we curated training data by removing videos containing on-screen clues. Table~\ref{tab:clue_detection} summarizes detection results. RepCount has the highest rate of affected videos (8.8\%), explaining higher reward hacking prevalence. In total, 105 videos (2.4\%) were removed from training data for our final experiments.

\begin{table}[t]
\centering
\small
\begin{tabular}{lccc}
\toprule
\textbf{Dataset} & \textbf{Total} & \textbf{With Clues} & \textbf{\%} \\
\midrule
OVR & 3,659 & 45 & 1.2\% \\
Custom Workout & 136 & 2 & 1.5\% \\
RepCount & 658 & 58 & 8.8\% \\
\midrule
\textbf{Total} & 4,453 & 105 & 2.4\% \\
\bottomrule
\end{tabular}
\caption{On-screen clue detection in training datasets. RepCount has highest rate (8.8\%) of videos with counters/timers; these were removed from training.}
\label{tab:clue_detection}
\end{table}

This exploitation pattern suggests current VLMs lack robust temporal reasoning capabilities, preferring text-based shortcuts over genuine motion understanding when available.

\section{Training Details}
\label{sec:training_details}

\subsection{Training Data Curation}

We aggregate training data from three public repetition counting datasets. Table~\ref{tab:data_raw} shows the raw dataset statistics before filtering.

\begin{table}[t]
\centering
\small
\begin{tabular}{lccc}
\toprule
\textbf{Dataset} & \textbf{Samples} & \textbf{Duration (s)} & \textbf{Count} \\
\midrule
OVR-Kinetics & 1,101 & 7.5 $\pm$ 1.6 & 11.2 $\pm$ 7.1 \\
RepCount & 787 & 30.5 $\pm$ 17.6 & 14.8 $\pm$ 14.8 \\
Countix & 1,870 & 9.4 $\pm$ 1.4 & 7.3 $\pm$ 7.0 \\
\midrule
\textbf{Total} & \textbf{3,758} & 13.3 $\pm$ 12.1 & 10.0 $\pm$ 9.7 \\
\bottomrule
\end{tabular}
\caption{Raw training data statistics before filtering.}
\label{tab:data_raw}
\end{table}

\subsection{Filtering Criteria}

We apply quality filters in two stages. First, temporal resolution and minimum count filters:

\paragraph{Nyquist Threshold.} Following the Nyquist-Shannon sampling theorem (see Section~\ref{sec:framerate}), we require $\geq$2 frames per repetition to reliably capture all action cycles. At our sampling rate of 5 fps, this means videos must have duration $\geq 0.4N$ seconds for $N$ repetitions.

\paragraph{Minimum Ground Truth.} We exclude samples with ground truth count $<$ 6, as very low counts provide insufficient training signal for learning temporal patterns and are often edge cases (partial repetitions, setup footage).

After the first stage of filtering, the intermediate training set contained 968 samples (540 OVR, 325 RepCount, 103 Custom Workout). However, post-submission analysis revealed two additional data quality issues necessitating further curation:

\paragraph{OVR Temporal Resolution.} OVR-Kinetics videos have a mean of only 3.3 frames per repetition at 5fps, with 46.4\% of samples having $<$3 frames/rep---insufficient for the model to visually distinguish individual cycles. By contrast, our custom annotations average 16.3 frames/rep (0\% below threshold) and RepCount averages 11.0 frames/rep (2.3\% below threshold). We removed OVR entirely.

\paragraph{GT=10 Mode Collapse.} Samples with ground-truth count of exactly 10 comprised 10.6\% of training data, creating a mode-collapse attractor. We removed all 37 such samples (12 from Custom Workout, 25 from RepCount).

The final curated training set contains \textbf{391 samples} (Table~\ref{tab:data_filtered}), a 60\% reduction that substantially improved all metrics (see Section 3.4).

\begin{table}[t]
\centering
\small
\begin{tabular}{lcccc}
\toprule
\textbf{Dataset} & \textbf{Before} & \textbf{After} & \textbf{Delta} \\
\midrule
OVR-Kinetics & 540 & 0 (removed) & $-$540 \\
Custom Workout & 103 & 91 & $-$12 (GT=10) \\
RepCount & 325 & 300 & $-$25 (GT=10) \\
\midrule
\textbf{Total} & \textbf{968} & \textbf{391} & \textbf{$-$59.6\%} \\
\bottomrule
\end{tabular}
\caption{Training data after full curation. OVR-Kinetics was removed entirely due to insufficient temporal resolution at 5fps. GT=10 samples were removed to prevent mode-collapse exploitation.}
\label{tab:data_filtered}
\end{table}

\subsection{Training Configuration}

We fine-tune Qwen3-VL-4B-Thinking using VERL \citep{verl2024} with the DAPO algorithm, which extends GRPO \citep{grpo2024} with three modifications: (1) no KL divergence penalty, (2) asymmetric clipping (clip-low=0.2, clip-high=0.28), and (3) token-level loss aggregation instead of sequence-level.

\paragraph{Hyperparameters.}
\begin{itemize}
\item \textbf{Hardware}: 4$\times$NVIDIA H100 GPUs
\item \textbf{Input}: 112 uniformly sampled frames at 360p resolution
\item \textbf{Batch size}: 128 (16 prompts $\times$ 4 rollouts $\times$ 2 gradient accumulation)
\item \textbf{Learning rate}: $1 \times 10^{-6}$ with cosine decay
\item \textbf{Rollout samples}: 4 per prompt (for advantage estimation)
\item \textbf{Training steps}: 144 (approximately 6 epochs over 391 samples, 24 steps/epoch)
\end{itemize}

\paragraph{Reward Function.}
We use a scale-invariant linear percentage reward:
\begin{equation}
r(\hat{c}, c_{\text{GT}}) = \max\left(0, 1 - \frac{|\hat{c} - c_{\text{GT}}|}{c_{\text{GT}}}\right)
\end{equation}
where $\hat{c}$ is the predicted count extracted from model output and $c_{\text{GT}}$ is the ground truth. This reward provides smooth gradients even when predictions are not exact, addressing the sparse gradient problem inherent to binary rewards (see Appendix~\ref{sec:reward_design} for detailed analysis of reward function design).

\paragraph{Checkpoint Selection.}
We select the checkpoint at step 96, where the training reward score plateaus. This was the only checkpoint evaluated on our benchmark from this run, removing any test-set leakage concern. Exact match accuracy can be misleading for checkpoint selection due to mode collapse (see Section~\ref{sec:metrics}).

\section{Frame Rate Ablation}
\label{sec:fps_ablation}

To empirically validate the Nyquist-based frame rate requirements discussed in Section~\ref{sec:framerate}, we conducted an ablation study varying sampling rate from 1 to 10 fps using Gemini 3 Flash. Table~\ref{tab:fps_ablation} shows that $R^2$ improves from 0.61 at 1 fps to 0.82 at 5 fps, with slight degradation at 10 fps (0.73). Exact accuracy also peaks at 5 fps (42.1\%).

\begin{table}[t]
\centering
\small
\begin{tabular}{lccc}
\toprule
\textbf{FPS} & \textbf{Exact (\%)} & \textbf{MAE} & \textbf{$R^2$} \\
\midrule
1 & 17.4 & 5.4 & 0.61 \\
2 & 36.8 & 4.0 & 0.67 \\
5 & \textbf{42.1} & \textbf{2.9} & \textbf{0.82} \\
10 & 40.5 & 3.5 & 0.73 \\
\bottomrule
\end{tabular}
\caption{FPS ablation on PushupBench using Gemini 3 Flash. Lower frame rates cause more videos to fall below the Nyquist threshold (F/Rep $<$ 2.0), degrading performance. $R^2$ peaks at 5 fps, suggesting optimal balance between temporal resolution and processing efficiency.}
\label{tab:fps_ablation}
\end{table}

This validates the theoretical analysis: at lower frame rates, more videos fall below the F/Rep $\geq$ 2.0 threshold required for reliable counting. The diminishing returns from 5 to 10 fps suggest that 5 fps provides sufficient temporal resolution for most exercise videos in our dataset, balancing accuracy against computational cost.

\section{Reward Function Design for GRPO}
\label{sec:reward_design}

GRPO computes advantages within each rollout group: $A_i = (r_i - \mu_g)/(\sigma_g + \epsilon)$. When all predictions receive identical rewards, $\sigma_g = 0$ and no gradient flows---pathological for exact-match rewards when the model cannot reliably hit the target.

We compared three reward functions: \textbf{Exact} ($r = \mathbf{1}[\hat{c} = c_{\text{GT}}]$), \textbf{Linear \%} ($r = \max(0, 1 - |\hat{c} - c_{\text{GT}}|/c_{\text{GT}})$), and \textbf{Stepped} (1.0 for exact, 0.5 for $\pm$1, 0.2 for $\pm$2, $-$0.3 otherwise).

\paragraph{Mode Collapse Risk.} Since GT=10 is the dataset mode (Figure~\ref{fig:gt_distribution}), always predicting ``10'' yields expected rewards of 0.106 (Exact), 0.651 (Linear \%), and 0.021 (Stepped). Linear \% is most vulnerable to this exploit. Table~\ref{tab:gt_prevalence} shows the reward each function assigns when predicting ``10'' for each ground truth value, weighted by training set prevalence.

\begin{table}[h]
\centering
\small
\begin{tabular}{@{}rrrrrr@{}}
\toprule
\textbf{GT} & \textbf{Count} & \textbf{Prev.} & \textbf{Linear\%} & \textbf{Exact} & \textbf{Stepped} \\
\midrule
10 & 103 & 10.6\% & 1.000 & 1.0 & +1.0 \\
6 & 94 & 9.7\% & 0.333 & 0.0 & $-$0.3 \\
11 & 77 & 8.0\% & 0.909 & 0.0 & +0.5 \\
9 & 68 & 7.0\% & 0.889 & 0.0 & +0.5 \\
12 & 67 & 6.9\% & 0.833 & 0.0 & +0.2 \\
8 & 54 & 5.6\% & 0.750 & 0.0 & +0.2 \\
16+ & 307 & 31.7\% & 0.3--0.6 & 0.0 & $-$0.3 \\
\midrule
\textbf{E[r]} & \textbf{968} & \textbf{100\%} & \textbf{0.651} & \textbf{0.106} & \textbf{+0.021} \\
\bottomrule
\end{tabular}
\caption{Reward for predicting ``10'' by ground truth in the initial 968-sample training set (GT$\geq$6). GT=10 dominance (10.6\%) makes mode collapse profitable for Linear \% (E[r]=0.651). This analysis motivated removing all GT=10 samples in our curated 391-sample set.}
\label{tab:gt_prevalence}
\end{table}

\begin{figure*}[t]
\centering
\includegraphics[width=\textwidth]{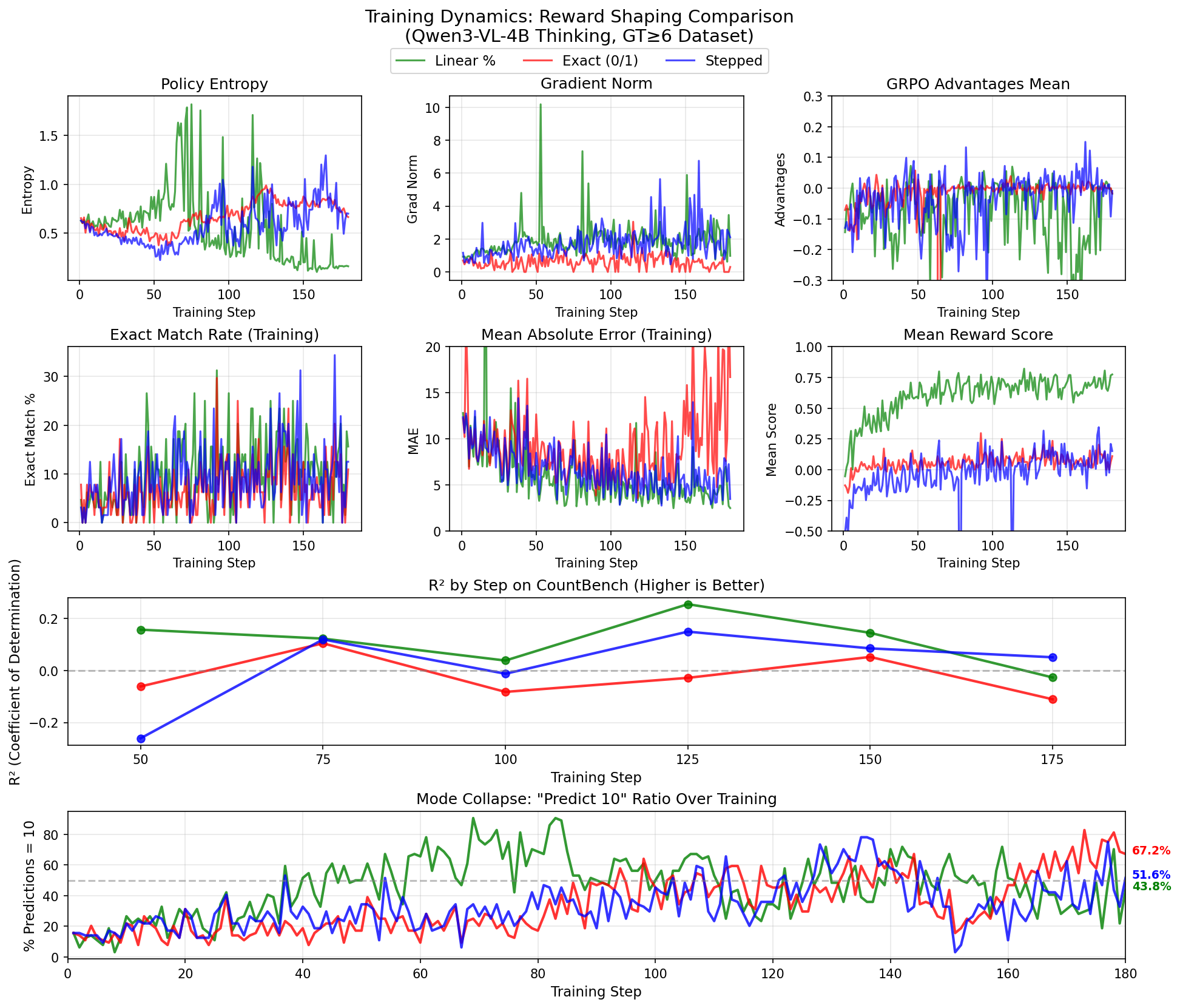}
\caption{Reward function ablation over 180 training steps (initial 968-sample experiment). We compare three rewards: \textbf{Exact} (binary 0/1), \textbf{Linear \%} (scale-invariant partial credit), and \textbf{Stepped} (discrete buckets with penalties). \textbf{Row 1}: Training diagnostics---entropy, gradient norm, GRPO advantages. \textbf{Row 2}: Task metrics---exact match, MAE, mean reward. \textbf{Row 3}: $R^2$ on PushupBench. \textbf{Row 4}: Mode collapse indicator---percentage predicting ``10''. \textbf{Key finding}: Linear \% achieves highest $R^2$ despite early mode collapse, because smooth gradients enable learning even during partial collapse. The final results in Table~\ref{tab:countbench} use the curated 391-sample dataset with Linear \% reward, which substantially improved over this initial experiment.}
\label{fig:training_dynamics}
\end{figure*}

\end{document}